\documentclass[11pt,letterpaper]{article}
\usepackage{naaclhlt2016}
\usepackage{times}
\usepackage{latexsym}
\usepackage{amsmath}
\usepackage{url}
\usepackage{graphicx}
\usepackage{amsfonts}

\usepackage[table]{xcolor}
\usepackage{array}
\usepackage{multirow}
\usepackage{tabularx}
\usepackage{multicol}
\usepackage{multirow}
\usepackage{diagbox}
\usepackage{amsmath}
\usepackage{booktabs}
\usepackage{float}
\restylefloat{table}
\usepackage[font=footnotesize,labelfont=bf]{caption}
\usepackage{enumitem}
\usepackage{makecell}
\usepackage{tikz}
\usetikzlibrary{calc}

\newcolumntype{x}[1]{>{\centering\arraybackslash}p{#1}}

\usepackage{hyperref}
\hypersetup{
  hidelinks 
}

\naaclfinalcopy
\def\naaclpaperid{***}

\title{Sequential Short-Text Classification with \\Recurrent and Convolutional Neural Networks}

\author{Ji Young Lee\thanks{\hspace{3mm}These authors contributed equally to this work.}\\
	    MIT\\
	    {\tt jjylee@mit.edu}
	   \And
	 Franck Dernoncourt\footnotemark[1]\\
   	MIT\\
   {\tt francky@mit.edu}}

\date{Dec 14, 2015}

\begin{document}

\maketitle
\begin{tikzpicture}[remember picture, overlay]
\node at ($(current page.north) + (-0in,-0.5in)$) {Accepted as a conference paper at NAACL 2016};
\end{tikzpicture}

\vspace{-0.5cm}

\begin{abstract}
\vspace{-0.2cm}Recent approaches based on artificial neural networks (ANNs) have shown promising results for short-text classification. 
However, many short texts occur in sequences (e.g., sentences in a document or utterances in a dialog), and most existing ANN-based systems do not leverage the preceding short texts when classifying a subsequent one.
In this work, we present a model based on recurrent neural networks and convolutional neural networks that incorporates the preceding short texts. 
Our model achieves state-of-the-art results on three different datasets for dialog act prediction.
\end{abstract}

\section{Introduction}

Short-text classification is an important task in many areas of natural language processing, including sentiment analysis, question answering, or dialog management. Many different approaches have been developed for short-text classification, such as using Support Vector Machines (SVMs) with rule-based features~\cite{silva2011symbolic}, combining SVMs with naive Bayes~\cite{wang2012baselines}, and building dependency trees with Conditional Random Fields~\cite{nakagawa2010dependency}. Several recent studies using ANNs have shown promising results, including convolutional neural networks (CNNs)~\cite{kim2014convolutional,blunsom2014convolutional,kalchbrenner2014convolutional} and recursive neural networks~\cite{socher2012semantic}.

Most ANN systems classify short texts in isolation, i.e., without considering preceding short texts.
However, short texts usually appear in sequence (e.g., sentences in a document or utterances in a dialog), therefore using information from preceding short texts may improve the classification accuracy.
Previous works on sequential short-text classification are mostly based on non-ANN approaches, such as Hidden Markov Models (HMMs) \cite{reithinger1997dialogue}, \cite{stolcke2000dialogue},
maximum entropy \cite{ang2005automatic}, and naive Bayes \cite{lendvai2007token}. 

\begin{figure*}[!ht]
  \centering
      \includegraphics[width=0.9\textwidth]{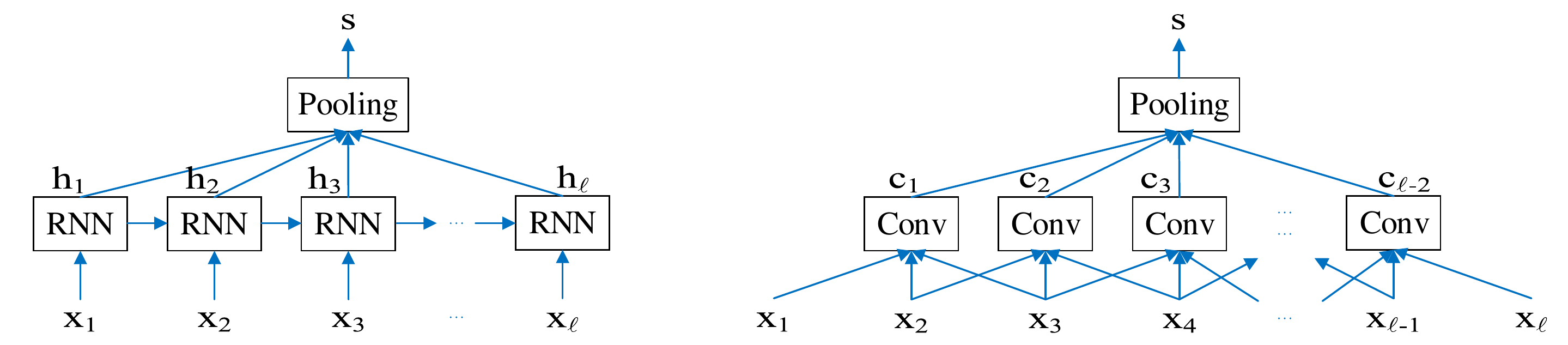}
\vspace{-0.0cm}
  \caption{RNN (left) and CNN (right) architectures for generating the vector representation $\mathbf{s}$ of a short text $\mathbf{x}_{1:\ell}$. For CNN, Conv refers to convolution operations, and the filter height $h=3$ is used in this figure.\vspace{-0.1cm}}
  \label{fig:sent2vec}
\end{figure*}

Inspired by the performance of ANN-based systems for non-sequential short-text classification, we introduce a model based on recurrent neural networks (RNNs) and CNNs for sequential short-text classification, and evaluate it on the dialog act classification task.
A dialog act characterizes an utterance in a dialog based on a combination of pragmatic, semantic, and syntactic criteria. Its accurate detection is useful for a range of applications, from speech recognition to automatic summarization~\cite{stolcke2000dialogue}. Our model achieves state-of-the-art results on three different datasets.

\section{Model} \label{sec:model}
\begin{figure*}[!ht]
  \centering
      \includegraphics[width=0.9\textwidth]{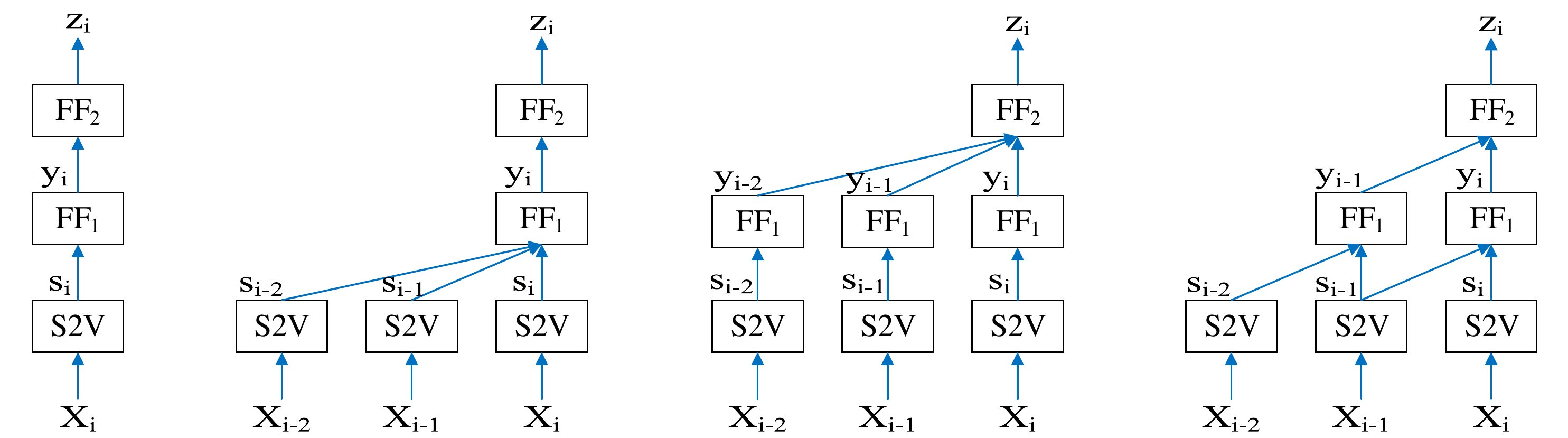}
\vspace{0.1cm}
  \caption{Four instances of the two-layer feedforward ANN used for predicting the probability distribution over the classes $\mathbf{z}_i$ for the $i^{th}$ short-text $\mathbf{X}_i$. S2V stands for short text to vector, which is the RNN/CNN architecture that generates $\mathbf{s}_i$ from $\mathbf{X}_i$. From left to right, the history sizes $(d_1, d_2)$ are $(0, 0), (2, 0), (0, 2)$ and $(1, 1)$. $(0, 0)$ corresponds to the non-sequential classification case.\vspace{-0.2cm}}
  \label{fig:sequential}
\end{figure*}

Our model comprises two parts. The first part generates a vector representation for each short text using either the RNN or CNN architecture, as discussed in Section~\ref{sec:representation} and Figure~\ref{fig:sent2vec}. The second part classifies the current short text based on the vector representations of the current as well as a few preceding short texts, as presented in Section~\ref{sec:sequential} and Figure~\ref{fig:sequential}.

We denote scalars with italic lowercases (e.g., $k,\, b_f$), vectors with bold lowercases (e.g., $\mathbf{s},\, \mathbf{x}_i$), and matrices with italic uppercases (e.g., $W_f$).
We use the colon notation $\mathbf{v}_{i:j}$ to denote the sequence of vectors $(\mathbf{v}_i, \mathbf{v}_{i+1}, \dotsc, \mathbf{v}_j)$.

\subsection{Short-text representation} \label{sec:representation}
A given short text of length $\ell$ is represented as the sequence of $m$-dimensional word vectors $\mathbf{x}_{1:\ell},$ which is used by the RNN or CNN model to produce the $n$-dimensional \textit{short-text representation} $\mathbf{s}.$   

\subsubsection{RNN-based short-text representation} \label{sec:rnn}
\noindent We use a variant of RNN called Long Short Term Memory (LSTM) \cite{hochreiter1997long}.
For the $t^{th}$ word in the short-text, an LSTM takes as input $\mathbf{x}_t, \mathbf{h}_{t-1}, \mathbf{c}_{t-1}$ and produces $\mathbf{h}_t, \mathbf{c}_t$ based on the following formulas:
\begin{align*}
\mathbf{i}_t &= \sigma(W_i \mathbf{x}_t + U_i \mathbf{h}_{t-1} + \mathbf{b}_i) \\
\mathbf{f}_t &= \sigma(W_f \mathbf{x}_t + U_f \mathbf{h}_{t-1} + \mathbf{b}_f) \\
\tilde{\mathbf{c}}_t &= \text{tanh}(W_c \mathbf{x}_t + U_c \mathbf{h}_{t-1} + \mathbf{b}_c) \\
\mathbf{c}_t &= \mathbf{f}_t \odot \mathbf{c}_{t-1} + \mathbf{i}_t \odot \tilde{\mathbf{c}}_t \\
\mathbf{o}_t &= \sigma(W_o \mathbf{x}_t + U_o \mathbf{h}_{t-1} + \mathbf{b}_o) \\
\mathbf{h}_t &= \mathbf{o}_t \odot \text{tanh}(\mathbf{c}_t) 
\end{align*}

\noindent where $W_j \in \mathbb{R}^{n \times m}, \; U_j \in \mathbb{R}^{n \times n}$ are weight matrices and $ \mathbf{b}_j \in \mathbb{R}^n$ are bias vectors, for $j \in \{i, f, c, o\}$.
The symbols $\sigma(\cdot)$ and tanh$(\cdot)$ refer to the element-wise sigmoid and hyperbolic tangent functions, and $\odot$ is the element-wise multiplication. $\mathbf{h}_0 = \mathbf{c}_0 = \mathbf{0}$.

In the pooling layer, the sequence of vectors $\mathbf{h}_{1:\ell}$
output from the RNN layer are combined into a single vector $\mathbf{s} \in \mathbb{R}^n$ that represents the short-text, using one of the following mechanisms: last, mean, and max pooling. Last pooling takes the last vector, i.e., $\mathbf{s} = \mathbf{h}_\ell,$ mean pooling averages all vectors, i.e., $\mathbf{s} = \tfrac{1}{\ell} \sum_{t=1}^\ell \mathbf{h}_t,$ and max pooling takes the element-wise maximum of $\mathbf{h}_{1:\ell}.$

\subsubsection{CNN-based short-text representation} \label{sec:cnn}
\noindent 
Using a \emph{filter} $W_f \in \mathbb{R}^{h \times m}$ of height $h,$  
a convolution operation on $h$ consecutive word vectors starting from $t^{th}$ word outputs the scalar \emph{feature} 
\vspace{-0.1cm}
\begin{align*}
c_t = \text{ReLU}(W_f \tiny{\, \bullet \,} X_{t:t+h-1} + b_f) 
\end{align*}
\vspace{-0.6cm}

\noindent where $X_{t:t+h-1} \in \mathbb{R}^{h \times m}$ is the matrix whose $i^{th}$ row is $\mathbf{x}_i \in \mathbb{R}^m,$
and $b_f \in \mathbb{R}$ is a bias. The symbol $\bullet$ refers to the dot product and ReLU$(\cdot)$ is the element-wise rectified linear unit function. 

We perform convolution operations with $n$ different filters, and denote the resulting features as $\mathbf{c}_t \in \mathbb{R}^n,$ each of whose dimensions comes from a distinct filter. 
Repeating the convolution operations for each window of $h$ consecutive words in the short-text, we obtain $\mathbf{c}_{1:\ell-h+1}.$
The short-text representation $\mathbf{s} \in \mathbb{R}^n$ is computed in the max pooling layer, as the element-wise maximum of $\mathbf{c}_{1:\ell-h+1}.$

\subsection{Sequential short-text classification} \label{sec:sequential}

Let $\mathbf{s}_i$ be the $n$-dimensional short-text representation given by the RNN or CNN architecture for the $i^{th}$ short text in the sequence.
The sequence $\mathbf{s}_{i-d_1-d_2 \, : \, i}$
is fed into a two-layer feedforward ANN that predicts the class for the $i^{th}$ short text. The hyperparameters $d_1, d_2$ are the history sizes used in the first and second layers, respectively.

The first layer takes as input
$\mathbf{s}_{i-d_1-d_2 \, : \, i}$ 
and outputs the sequence
$\mathbf{y}_{i-d_2\,:\,i}$ defined as
\vspace{-0.2cm}
\begin{align*}
\mathbf{y}_j = \text{tanh} \left( \sum_{d = 0}^{d_1} W_{-d} \, \mathbf{s}_{j-d} + \mathbf{b}_1 \right), \; \forall j \in [i-d_2, i]
\end{align*}\vspace{-0.2cm}

\noindent where $W_0, W_{-1}, W_{-2} \in \mathbb{R}^{k \times n}$ are the weight matrices, 
$\mathbf{b}_1 \in \mathbb{R}^k$ is the bias vector,
$\mathbf{y}_j \in \mathbb{R}^k$ is the \textit{class representation}, 
and $k$ is the number of classes for the classification task.

Similarly, the second layer takes as input the sequence of class representations 
$\mathbf{y}_{i-d_2:i}$  
 and outputs $\mathbf{z}_i \in \mathbb{R}^k$:
 \vspace{-0.3cm}
 \begin{align*}
\mathbf{z}_i = \text{softmax} \left( \sum_{j = 0}^{d_2} W_{-j} \, \mathbf{y}_{i-j} + \mathbf{b}_2 \right) 
\end{align*}\vspace{-0.3cm}

\noindent where $U_0, U_{-1}, U_{-2} \in \mathbb{R}^{k \times k}$ and $\mathbf{b}_2 \in \mathbb{R}^k$ are the weight matrices and bias vector.

The final output $\mathbf{z}_i$ represents the probability distribution over the set of $k$ classes for the $i^{th}$ short-text: the $j^{th}$ element of $\mathbf{z}_i$ corresponds to the probability that the $i^{th}$ short-text belongs to the $j^{th}$ class.

\section{Datasets and Experimental Setup}

\subsection{Datasets}
We evaluate our model on the dialog act classification task using the following datasets: 
\begin{itemize}[leftmargin=*]
 \setlength\itemsep{-0.1em}
\item DSTC 4: Dialog State Tracking Challenge 4~\protect\cite{DSTC4handbook,DSTC4}. 

\item MRDA: ICSI Meeting Recorder Dialog Act Corpus~\protect\cite{janin2003icsi,shriberg2004icsi}. 
The 5 classes are introduced in~\protect\cite{ang2005automatic}. 

\item SwDA: Switchboard Dialog Act Corpus~\protect\cite{jurafsky1997switchboard}. 
\end{itemize}
For MRDA, we use the train/validation/test splits provided with the datasets. For DSTC 4 and SwDA, only the train/test splits are provided.\footnote{All train/validation/test splits can be found at \url{https://github.com/Franck-Dernoncourt/naacl2016}} 
Table~\ref{tab:datasets} presents statistics on the datasets.

\vspace{-0.2cm}
\begin{table} [H]
\footnotesize
\centering
\setlength\tabcolsep{4.0pt}
\setlength{\extrarowheight}{3pt}
\setlength{\arraycolsep}{5pt}
\begin{tabular}{|l|c|c|c|c|c|}
\hline
\textbf{Dataset} & \textbf{$|C|$} 	& \textbf{$|V|$} 	& Train & Validation & Test \\
\hline
\text{DSTC 4}	& 89	&	6k	& 24 (21k) 	& 5 (5k)	 	& 6 (6k)\\
\text{MRDA}		& 5		&	12k		& 51 (78k) 	& 11 (16k)	& 11 (15k)	 \\
\text{SwDA}		& 43	&	20k		& 1003 (193k) 	& 112 (23k)	& 19 (5k) \\
\hline
\end{tabular}
\caption{Dataset overview.   
$|C|$ is the number of classes, $|V|$ the vocabulary size. For the train, validation and test sets, we indicate the number of dialogs (i.e., sequences) followed by the number of utterances (i.e., short texts) in parenthesis.} \label{tab:datasets}
\end{table}

\vspace{-0.5cm}

\subsection{Training} \label{sec:training}
The model is trained to minimize the negative log-likelihood of predicting the correct dialog acts of the utterances in the train set, using stochastic gradient descent with the Adadelta update rule~\cite{zeiler2012adadelta}.
At each gradient descent step, weight matrices, bias vectors, and word vectors are updated.
For regularization, dropout is applied after the pooling layer, and early stopping is used on the validation set with a patience of 10 epochs.

\section{Results and Discussion}

To find effective hyperparameters, we varied one hyperparameter at a time while keeping the other ones fixed. Table \ref{tab:hyperparameter} presents our hyperparameter choices.

\vspace{-0.1cm}
\begin{table} [ht]
\footnotesize
\centering
\setlength{\extrarowheight}{3pt}
\setlength{\arraycolsep}{5pt}
\begin{tabular}{|l|c|c|}
\hline
\textbf{Hyperparameter} & \textbf{Choice} 	& \textbf{Experiment Range} \\
\hline
\text{LSTM output dim. $(n)$}		& 100				& 50 -- 1000 \\
\text{LSTM pooling}					& \text{max}		& \text{max, mean, last} \\
\text{LSTM direction}				& \text{unidir.}	& \text{unidir., bidir.}\\	
\text{CNN num. of filters $(n)$}	& 500				& 50 -- 1000 \\
\text{CNN filter height $(h)$}		& 3					& 1 -- 10\\
\text{Dropout rate}					& 0.5				& 0 -- 1\\
\text{Word vector dim. $(m)$}		& 200, 300			& 25 -- 300\\
\hline
\end{tabular}
\caption{Experiments ranges and choices of hyperparameters. Unidir refers to the regular RNNs presented in Section~\ref{sec:rnn}, and bidir refers to bidirectional RNNs introduced in ~\protect\cite{schuster1997bidirectional}.}\label{tab:hyperparameter}
\end{table}

\vspace{-0.1cm}

\begin{table*} [ht]
\scriptsize
\centering
\setlength{\extrarowheight}{3pt}
\setlength{\arraycolsep}{5pt}
\begin{tabular}{|x{3.5em}x{1.5em}| ccc | ccc|}
\hline

\multicolumn{2}{|c|}{ \multirow{2}{*}{ \hspace{-0.07cm}\diagbox[width=8.1em,height=3.15em]{ $d_1$}{$d_2$ }} } 
																& \multicolumn{3}{c|}{\textbf{LSTM}}    									& \multicolumn{3}{c|}{\textbf{CNN}}\\ 

\multicolumn{2}{|x{5em}|}{} 					& 0  				& 1					& 2    			& 0  				& 1					& 2    \\
\hline
\multirow{3}{*}{\textbf{DSTC4}}	
				&0 				& 63.1\, (62.4, 63.6) 	& 65.7\, (65.6, 65.7) 	& 64.7\, (63.9, 65.3) 	& 64.1\, (63.5, 65.2) 	& 65.4\, (64.7, 66.6) 	& 65.1\, (63.2, 65.9) \\
				&1 		& \textbf{65.8}\, (65.5, 66.1) 	& 65.7\, (65.3, 66.1) 	& 64.8\, (64.6, 65.1) 	& 65.3\, (64.1, 65.9) 	& 65.1\, (62.1, 66.2) 	& 64.9\, (64.4, 65.6) \\
				&2 				& 65.7\, (65.0, 66.2) 	& 65.5\, (64.4, 66.1) 	& 64.9\, (64.6, 65.2) 	& 65.7\, (64.9, 66.3) 	& \textbf{65.8}\, (65.2, 66.1) 	& 65.4\, (64.5, 66.0) \\
\hline
\multirow{3}{*}{\textbf{MRDA}}	
				&0				& 82.8\, (82.4, 83.1) 	& 83.2\, (82.9, 83.4) 	& 82.9\, (82.4, 83.4) 	& 83.2\, (83.0, 83.4) 	& 83.5\, (82.9, 84.0) 	& 83.8\, (83.4, 84.2) \\
				&1				& 83.2\, (82.6, 83.7) 	& 83.8\, (83.5, 84.4) 	& 83.6\, (83.2, 83.8) 	& \textbf{84.6}\, (84.5, 84.9) 	& \textbf{84.6}\, (84.4, 84.8) 	& 84.1\, (83.8, 84.4) \\
				&2		& \textbf{84.1}\, (83.5, 84.4) 	& 83.9\, (83.4, 84.7) 	& 83.3\, (82.6, 84.2) 	& 84.4\, (84.1, 84.8) 	& \textbf{84.6}\, (84.5, 84.7) 	& 84.4\, (84.2, 84.7) \\
\hline
\multirow{3}{*}{\textbf{SwDA}}	
				&0				& 66.3\, (65.1, 68.0) 	& 67.9\, (66.3, 68.6) 	& 67.8\, (66.7, 69.0) 	& 67.0\, (65.3, 68.7) 	& 69.1\, (68.5, 70.0) 	& 69.7\, (69.2, 70.9) \\
				&1				& 68.4\, (67.8, 68.8) 	& 67.8\, (65.5, 68.9) 	& 67.3\, (65.5, 69.5) 	& 69.9\, (69.1, 70.9) 	& 69.8\, (69.3, 70.6) 	& 69.9\, (68.8, 70.6) \\ 
				&2		& \textbf{69.5}\, (68.9, 70.2) 	& 67.9\, (66.5, 69.4) 	& 67.7\, (66.9, 68.9) 	& \textbf{71.4}\, (70.4, 73.1) 	& 71.1\, (70.2, 72.1) 	& 70.9\, (69.7, 71.7) \\ 
\hline
\end{tabular}
\renewcommand\thetable{3}
\caption{Accuracy (\%) on different architectures and history sizes $d_1, d_2$. For each setting, we report average (minimum, maximum) computed on 5 runs. 
Sequential classification ($d_1 + d_2 > 0$) outperforms non-sequential classification ($d_1 = d_2 = 0$).
Overall, the CNN model outperformed the LSTM model for all datasets, albeit by a small margin except for SwDA. We also tried a variant of the LSTM model, gated recurrent units~\protect\cite{cho2014properties}, but the results were generally lower than LSTM.\vspace{-0.2cm}} \label{tab:results}
\end{table*}

We initialized the word vectors with the 300-dimensional word vectors pretrained with word2vec on Google News
~\cite{mikolov2013efficient,mikolov2013distributed} for DSTC~4, and the 200-dimensional word vectors pretrained with GloVe on Twitter
~\cite{pennington2014glove} for MRDA and SwDA, as these choices yielded the best results among all publicly available word2vec, GloVe, SENNA~\cite{collobert2011deep,collobert2011natural} and RNNLM~\cite{mikolov2011rnnlm} word vectors.

The effects of the history sizes $d_1$ and $d_2$ for the short-text and the class representations, respectively,
are presented in Table \ref{tab:results} for both the LSTM and CNN models. In both models, increasing $d_1$  while keeping $d_2 = 0$
improved the performances by 1.3-4.2 percentage points. Conversely, increasing $d_2$ while keeping $d_1 = 0$ yielded better results, but the performance increase was less pronounced:
incorporating sequential information at the short-text representation level was more effective than at the class representation level.

Using sequential information at both the short-text representation level and the class representation level does not help in most cases and may even lower the performances.
We hypothesize that short-text representations contain richer and more general information than class representations due to their larger dimension. Class representations may not convey any additional information over short-text representations, and are more likely to propagate errors from previous misclassifications.

Table~\ref{tab:result-comparisons} compares our results with the state-of-the-art. 
Overall, our model shows competitive results, while requiring no human-engineered features.
Rigorous comparisons are challenging to draw, as many important details such as text preprocessing and train/valid/test split may vary, and many studies fail to perform several runs despite the randomness in some parts of the training process, such as weight initialization.

\vspace{-0.1cm}
\begin{table} [H]
\footnotesize
\centering
\setlength{\extrarowheight}{3pt}
\setlength{\arraycolsep}{5pt}
\begin{tabular}{|l|c|c|c|c|c|c|}
\hline
\textbf{Model} & DSTC~4 	& MRDA 	& SwDA\\
\hline
CNN			& 65.5			&	\textbf{84.6}	& \textbf{73.1} \\ 
LSTM		& \textbf{66.2}			&	84.3	& 69.6 \\ 
Majority class		& 25.8			&	59.1	& 33.7 \\
SVM		& 57.0		&	 --		& --	\\
Graphical model	& --		&	 81.3		& --	\\
Naive Bayes & --			&	 82.0		& --	\\
HMM & --			&	 --		& 71.0	\\
Memory-based Learning & --			&	 --		& 72.3	\\
Interlabeler agreement & --			&	 --		& 84.0	\\
\hline
\end{tabular}
\caption{Accuracy (\%) of our models and other methods from the literature. 
The majority class model predicts the most frequent class. 
SVM: \protect\cite{dernoncourt2016adobe}.
Graphical model: \protect\cite{ji2006backoff}.
Naive Bayes: \protect\cite{lendvai2007token}.
HMM: \protect\cite{stolcke2000dialogue}.
Memory-based Learning: \protect\cite{rotaru2002dialog}. 
All five models use features derived from transcribed words, as well as previous predicted dialog acts except for Naive Bayes.
The interlabeler agreement could be obtained only for SwDA.
For the CNN and LSTM models, the presented results are the test set accuracy of the run with the highest accuracy on the validation set.\vspace{-0.1cm}
} \label{tab:result-comparisons}
\end{table}

\vspace{-0.3cm}

\section{Conclusion}

In this article we have presented an ANN-based approach to sequential short-text classification. We demonstrate that adding sequential information improves the quality of the predictions, and the performance depends on what sequential information is used in the model.
Our model achieves state-of-the-art results on three different datasets for dialog act prediction.

\newpage
\bibliography{naaclhlt2016}
\bibliographystyle{naaclhlt2016}

\end{document}